\documentclass[conference]{IEEEtran}
\usepackage{cite}
\usepackage{amsmath,amssymb,amsfonts}
\usepackage{graphicx}
\usepackage{booktabs}

\def\BibTeX{{\rm B\kern-.05em{\sc i\kern-.025em b}\kern-.08em
    T\kern-.1667em\lower.7ex\hbox{E}\kern-.125emX}}

\begin{document}

\title{Parser-Free VLM Verification for Federated Weakly Supervised Video Anomaly Detection}

\author{
\IEEEauthorblockN{
Sébastien Thuau\textsuperscript{1,2},
Amira Gran\textsuperscript{1},
Siba Haidar\textsuperscript{1},
Rachid Chelouah\textsuperscript{2}
}
\IEEEauthorblockA{
\textsuperscript{1}\textit{esieaLab, ESIEA}, Paris, France \quad
\textsuperscript{2}\textit{ETIS Laboratory, CNRS, UMR8051, CY Cergy Paris University}, Paris, France
}
\IEEEauthorblockA{
sebastien.thuau@cyu.fr,\ amira.gran@et.esiea.fr,\ siba.haidar@esiea.fr,\ rachid.chelouah@cyu.fr
}
}

\maketitle

\begin{abstract}
How can vision-language models help video anomaly detection (VAD) when surveillance data remain distributed, weakly labeled, and resource-constrained? Most weakly supervised VAD methods assume centralized training; recent VLM-based extensions further rely on dense inference, generated explanations, or additional adaptation. We introduce a lightweight federated MIL--VLM cascade in which only a compact MIL scorer is trained across clients, while a frozen VLM verifies high-scoring suspect segments post hoc. We study two VLM feedback interfaces: parsed text-generation decisions and a logit-based interface that extracts a continuous anomaly score from next-token \emph{Yes}/\emph{No} probabilities. Experiments on UCF-Crime with InternVL3.5-2B and Qwen3-VL-2B-Instruct show that text-generation verification can improve frame-level AUC after diagnostic temporal post-processing, but remains sensitive to prompts, parsers, model choice, and smoothing. In contrast, the logit interface provides a fixed parser-free signal that improves both frame-level AUC and frame-level AP over the MIL baseline across both VLMs, without temporal post-processing in its main configuration. Since suspect segments are updated independently once available, next-token logit feedback provides a simple segment-local alternative to text-generation verification.
\end{abstract}


\section{Introduction}

Video anomaly detection (VAD) aims to identify rare and safety-critical events in long surveillance videos. In realistic deployments, surveillance data are distributed across cameras or sites, raw footage is difficult to centralize because of privacy and bandwidth constraints, and client devices may have limited computational power. At the same time, training labels are typically available only at the video level, while evaluation requires frame- or segment-level localization. These constraints motivate federated weakly supervised VAD methods that can operate with compact trainable components.

Weakly supervised VAD naturally fits a multiple-instance learning (MIL) setting: a video-level label supervises a bag of temporal segments, and the model must infer which segments explain the label. This formulation is compatible with lightweight federated learning when the trainable component operates on frozen visual features. However, MIL scores are inferred from coarse supervision alone, so high-scoring segments do not always correspond to semantically abnormal events. Visually salient but normal segments can therefore be ranked as anomalous without any explicit semantic verification.

Vision-language models (VLMs) offer a natural way to verify whether a suspect segment is semantically abnormal by relating visual evidence to language-level concepts. Existing VLM-based VAD methods, however, often rely on dense inference, learned prompts, generated explanations, or chain-of-thought reasoning, which increases computation and introduces dependencies on prompt wording and text parsing. We instead use frozen off-the-shelf VLMs only as selective verifiers for segments proposed by a lightweight federated MIL scorer. The VLM is not trained, synchronized, or aggregated across clients; in our experiments it is executed offline in a centralized evaluation setting for reproducible comparison, while the same frozen verification step can be deployed locally or on-premise when hardware permits.

This raises the central question of this paper: \emph{how should feedback from a frozen VLM be represented so that semantic verification remains efficient, parser-free, and compatible with causal segment-level scoring?}

The contributions of this paper are:
\begin{itemize}
    \item We propose a lightweight federated MIL--VLM cascade in which only a compact MIL head is trained across physical CPU-only clients, while a frozen VLM verifies selected suspect segments post hoc.
    \item We show that text-generation VLM verification is sensitive to prompting, parsing, model choice, and temporal post-processing.
    \item We propose a logit-based VLM interface that extracts a continuous anomaly score from next-token \emph{Yes}/\emph{No} probabilities, avoids keyword parsing, preserves ranking information, and remains causal at the segment-scoring level in its main configuration.
\end{itemize}

\section{Related Work}

\subsection{Weakly Supervised Video Anomaly Detection}

Weakly supervised VAD uses only video-level labels for training while requiring frame- or segment-level localization at evaluation. Sultani et al.~\cite{sultani2018real} introduced the MIL formulation for WSVAD on the UCF-Crime benchmark, treating videos as bags of temporal segments with sparsity and smoothness constraints. Subsequent methods improved weakly supervised VAD through stronger temporal modeling, feature magnitude learning, memory mechanisms, and bias reduction~\cite{tian2021rtfm,chen2023mgfn,zhou2023urdmu,lv2023unbiased}.

Recent work also incorporates vision-language priors into weakly supervised VAD. VadCLIP~\cite{wu2024vadclip} adapts CLIP~\cite{radford2021clip} to anomaly detection with visual and language-alignment branches, while prompt-enhanced MIL approaches such as PE-MIL~\cite{chen2024pemil} use textual semantics to improve anomaly localization. These methods improve weakly supervised VAD but assume centralized training and do not address client heterogeneity or edge-oriented constraints. In contrast, our work focuses on a lightweight federated MIL scorer that queries a frozen VLM selectively, rather than treating the VLM as a dense detector.

\subsection{Federated Video Anomaly Detection}

Federated learning~\cite{mcmahan2017fedavg} trains a shared model across clients without centralizing data, but client data are often non-IID and heterogeneous in practical surveillance settings. Recent VAD-specific federated approaches have begun to address these issues. FedVAD introduces a federated VAD framework that distills knowledge from GPT-generated semantic descriptions to improve learning under data heterogeneity~\cite{qi2024fedvad}. FedWSAD extends weakly supervised VAD to a federated setting through multimodal prompt learning with global and local contexts~\cite{wang2025fedwsad}. Su et al. propose a mixture of local-to-global experts for federated weakly supervised VAD to handle non-IID client distributions~\cite{su2025fedwsvad}.

These works federate either the detector itself or auxiliary mechanisms such as prompt learning, semantic distillation, or expert aggregation. Our work differs in keeping the VLM frozen and using it as a post-hoc selective verifier; the federated component is only the lightweight MIL scorer. Our main question concerns the interface between the federated scorer and the frozen VLM.

\subsection{Vision-Language Models for Video Anomaly Detection}

Recent work leverages vision-language models to improve the semantic interpretability of VAD. LAVAD~\cite{zanella2024lavad} uses large language models for training-free anomaly detection from video descriptions, while AnomalyRuler~\cite{yang2024anomalyruler} introduces rule-based LLM reasoning from few normal references. VERA~\cite{ye2025vera} learns guiding questions that elicit anomaly reasoning in frozen VLMs without modifying model parameters, and Vad-R1~\cite{huang2025vadr1} studies anomaly reasoning through perception-to-cognition chain-of-thought prompting. These methods show the promise of VLMs for semantic anomaly understanding, but they mainly rely on generated text, prompt design, or reasoning traces.

A second line of work improves efficiency by focusing VLM computation on informative regions. Holmes-VAU~\cite{zhang2025holmes} concentrates VLM reasoning on anomaly-rich regions through anomaly-focused temporal sampling, but does not address weakly supervised federated settings. SlowFastVAD~\cite{ding2025slowfastvad} routes selected ambiguous segments from a fast detector to a slower RAG-enhanced VLM before fusing scores, making it closest in spirit to our selective verification approach. It differs from our work in being centralized, in using retrieval-augmented reasoning rather than logit-based verification, and in not analyzing causal segment-level scoring. Next-token likelihoods are standard in language model evaluation, but their use as a continuous verification interface for weakly supervised VAD remains, to our knowledge, unexplored.

\section{Methodology}

\subsection{Problem Setting and Pipeline Overview}
\label{sec:method_overview}

We address weakly supervised VAD: each video is divided into $T$ temporal segments, training supervision is available only at the video level, and segment scores are expanded to frame-level scores for evaluation.

\begin{figure}
    \centering
    \includegraphics[width=\linewidth]{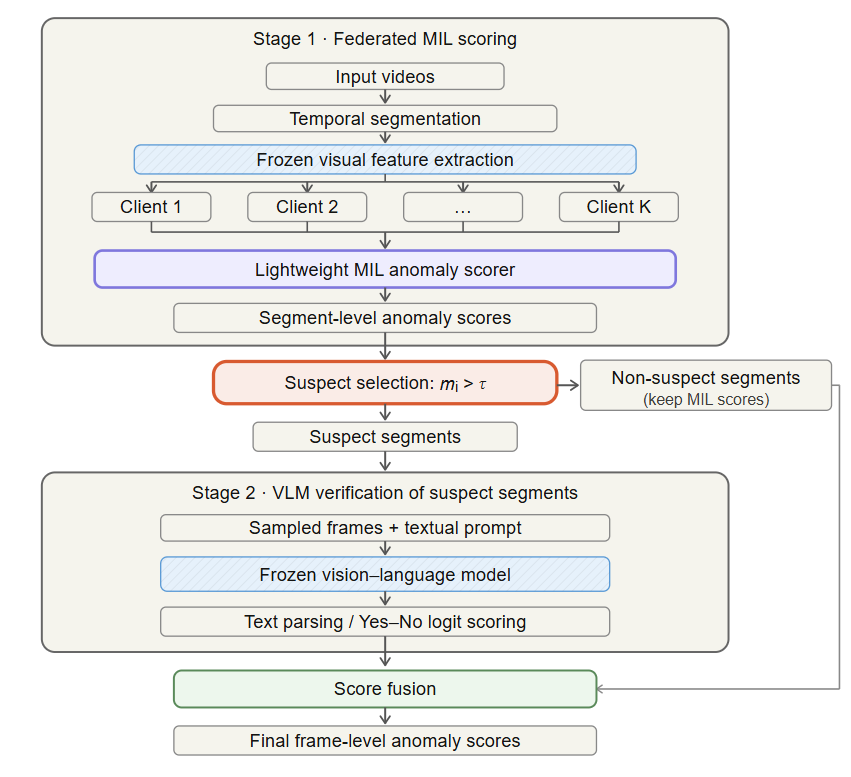}
    \caption{Two-stage MIL--VLM cascade. The federated MIL scorer routes suspect segments to a frozen VLM; non-suspect segments keep their MIL scores.}
    \label{fig:pipeline}
\end{figure}

Our method follows the two-stage pipeline shown in Fig.~\ref{fig:pipeline}. Stage~1 trains a lightweight federated multiple-instance learning (MIL) scorer on frozen visual features. For each segment $i$, it outputs an anomaly score $m_i$, producing a dense temporal anomaly profile. Stage~2 uses a frozen vision-language model (VLM) only on suspect segments selected from this MIL profile, making the VLM a selective semantic verifier rather than a dense detector.

Federated learning is used only to train the MIL scorer. The VLM is never trained, updated, or aggregated across clients; it is used post hoc to rescore selected suspect segments. Thus, verification can be run centrally for offline evaluation, as in our experiments, or locally/on-premise when client hardware permits.

\subsection{Federated MIL Anomaly Scorer}
\label{sec:federated_mil}

Stage~1 learns a compact MIL anomaly scorer from video-level labels. Each segment is represented by a frozen CLIP ViT-B/16 feature $\mathbf{x}_i \in \mathbb{R}^{512}$~\cite{radford2021clip}. The CLIP encoder is fixed; only the MIL head is trained in a federated manner.

The anomaly head is an MLP with layer widths $512 \rightarrow 128 \rightarrow 32 \rightarrow 1$, ReLU activations, high dropout ($p=0.6$), and a final sigmoid. It outputs a segment-level anomaly score $m_i \in [0,1]$ and contains 69.8K trainable parameters.

The video-level prediction is obtained by averaging the top-$k$ segment scores, with $k=3$ for all experiments. The model is trained with video-level binary cross-entropy plus sparsity and temporal smoothness regularization~\cite{sultani2018real}. The MIL head is trained with FedAvg~\cite{mcmahan2017fedavg}; to emulate client heterogeneity, training videos are partitioned across clients with a Dirichlet-based label allocation over video-level classes.

\subsection{Selective VLM Verification}
\label{sec:vlm_verification}

Given MIL scores $\{m_i\}_{i=1}^{T}$, a segment $i$ is selected for VLM verification when $m_i > \tau$, where $\tau$ is fixed in Sec.~\ref{sec:experimental_setup}. For each suspect segment, we sample frames and pair them with a textual prompt. Non-suspect segments are not sent to the VLM and keep their MIL scores. In the reported experiments, VLM calls are run offline after federated training; no VLM parameters, gradients, generated labels, or generated descriptions are exchanged during federated optimization.

The design question is how to convert frozen VLM output into an anomaly signal. We study two interfaces: generated text followed by parsing, and direct next-token \emph{Yes}/\emph{No} logit extraction.

\subsubsection{Text-Generation Interface}
In the text-generation regime, the VLM response is parsed using case-insensitive keyword matching. Responses containing ``anomalous'' or ``abnormal'' confirm the MIL score; responses containing ``normal'' but not ``abnormal'' or ``anomalous'' are parsed as normal and attenuate the score according to Sec.~\ref{sec:score_fusion}. Unparsed responses leave the segment unchanged. This rule does not handle negations and may misclassify some chain-of-thought outputs, as discussed in Sec.~\ref{sec:prompt_exploration}.

We also evaluate a diagnostic score-output prompt, R13, which asks for a 0--10 anomaly score. When valid, we convert it to $q_i=\mathrm{score}_i/10$; when outputs collapse to a few discrete values, we treat the mapping as model-specific parser adaptation, reported only as diagnostic.

\subsubsection{Logit-Based Interface}
Text generation requires decoding and keyword parsing. To avoid this dependency, we use the fixed normality prompt: ``Is this surveillance scene normal? Answer only Yes or No.'' We extract next-token logits for \emph{Yes} and \emph{No}; since the prompt asks about normality, \emph{No} corresponds to an anomaly. Candidate token IDs are fixed before evaluation for each VLM tokenizer. We verify under each model's inference template that these candidates are single-token continuations and use the restricted two-token softmax as a ranking signal rather than as a calibrated probability.

Let $\ell_{\mathrm{yes},i}$ and $\ell_{\mathrm{no},i}$ denote the logits for suspect segment $i$. We apply a softmax restricted to these two logits and use the probability assigned to \emph{No}:
\[
p^{\mathrm{VLM}}_i
=
\frac{\exp(\ell_{\mathrm{no},i})}
{\exp(\ell_{\mathrm{yes},i}) + \exp(\ell_{\mathrm{no},i})}.
\]
This gives a continuous, parser-free anomaly score. In the main $\lambda=1$ configuration without temporal smoothing, each suspect segment is updated independently and the verifier is causal at the segment-scoring level.

\subsection{Score Substitution, Fusion, and Temporal Post-Processing}
\label{sec:score_fusion}

After VLM verification, non-suspect segments keep their MIL score $m_i$ and only suspect segments can be updated. For text generation, the default rule attenuates segments parsed as normal, $s_i=0.1m_i$, while confirmed anomalies and unparsed responses remain unchanged.

In the exploratory post-processing sweep, we also test linear fusion
\[
s_i=\lambda q_i+(1-\lambda)m_i,
\]
where $q_i=0$ for normal-corrected segments, $q_i=1$ for anomaly-confirmed segments, and $q_i=\mathrm{score}_i/10$ for valid score-output prompts.

For logit verification, the VLM directly provides $p_i^{\mathrm{VLM}}$ and suspect segments are updated as
\[
s_i = \lambda p_i^{\mathrm{VLM}} + (1-\lambda)m_i .
\]
The main logit configuration uses $\lambda=1$, substituting suspect MIL scores with VLM anomaly probabilities.

For text-generation post-processing, we smooth correction signals with EMA and local majority voting, either on the full $T=32$ timeline or on the compressed suspect sequence. Offline smoothing uses bidirectional EMA or centered neighborhoods; causal smoothing uses forward-only EMA and past-only neighborhoods. The sweep covers attenuation, linear fusion with $\lambda\in\{0.1,0.3,0.5,0.7,0.9\}$, EMA $\beta\in\{0.1,0.3,0.5,0.7\}$, majority windows $w\in\{1,2,3,5,7\}$, and both smoothing domains. The main logit configuration uses no temporal smoothing.

\section{Experiments and Results}

\subsection{Experimental Setup and Dataset}
\label{sec:experimental_setup}

We evaluate our approach on UCF-Crime, a surveillance video anomaly detection benchmark containing normal videos and 13 anomaly categories~\cite{sultani2018real}. Video-level labels are used for training, while frame-level annotations are used only for evaluation. The test set contains 290 videos with frame-level anomaly annotations.

Each video is uniformly divided into $T=32$ temporal segments. Segment scores are expanded to frame-level scores by assigning each frame the score of its corresponding segment. We report frame-level AUC (F-AUC) and frame-level AP (F-AP) over all test frames.

For the federated Stage~1 MIL scorer, we use $K=4$ physical CPU-only client machines. CLIP features are pre-extracted, and each client trains only the MIL head on frozen segment features. Training videos are partitioned across clients using a Dirichlet-based allocation over video-level labels with concentration parameter $\alpha=0.5$. FedAvg is run for $R=30$ communication rounds with $E=1$ local epoch per round.

For Stage~2, suspect segments are selected with a fixed threshold $\tau=0.5$, shared across all VLMs and prompts. For reproducibility, all VLM verification experiments are performed offline in a centralized evaluation environment after federated MIL training. This choice standardizes hardware and model access across VLMs and prompts. It should not be interpreted as a fully privacy-preserving inference protocol: if frames are sent to a central verifier, privacy depends on the deployment setting. However, because the verifier is frozen and segment-wise, the same procedure can be executed locally or on-premise by clients with sufficient VLM inference capability.

With $\tau=0.5$, the pipeline routes $8.09 \pm 0.41$ segments per video to the VLM on average, corresponding to $25.3\% \pm 1.3\%$ of the 32 segments and a $3.95\times$ reduction in VLM calls compared with dense verification. Dense zero-shot VLM scoring is reported as a baseline in Table~\ref{tab:stage1_backbones}. Averaged across the five Stage~1 checkpoints used for cascade evaluation, introduced in Sec.~\ref{sec:stage1_results}, $51.7\%$ of test videos have no suspect segment and skip the VLM stage entirely.

For each suspect segment, we uniformly sample two frames resized to $448 \times 448$ as a fixed cost-context tradeoff. We evaluate two frozen VLMs: InternVL3.5-2B~\cite{wang2025internvl35} and Qwen3-VL-2B-Instruct~\cite{bai2025qwen3vl}. Text-generation uses deterministic decoding, while the logit regime extracts the next-token \emph{Yes}/\emph{No} logits from a single generated token. The constants $k=3$, $\tau=0.5$, two-frame VLM input, and attenuation factor $0.1$ are fixed before the main VLM comparison. The text-generation strategy sweep in Sec.~\ref{sec:prompt_exploration} is diagnostic and should not be interpreted as a validation protocol for prompt selection.

\subsection{Stage-1 Backbone Screening and Checkpoint Selection}
\label{sec:stage1_results}

We first screen Stage~1 backbones before VLM verification. The goal is to select a practical representation for the MIL--VLM cascade, not to provide an exhaustive backbone benchmark. We compare CLIP+MLP with compact 3D CNN baselines trained from low-resolution clips; 3D CNN-64 and 3D CNN-112 denote the same 3D CNN trained on $64 \times 64$ and $112 \times 112$ clips.

Table~\ref{tab:stage1_backbones} also reports dense zero-shot VLM baselines, where the VLM is queried on all 32 segments without MIL routing. Basic uses the minimal binary prompt ``Is there an anomaly in this surveillance scene? Answer only Yes or No.'' VERA denotes the VERA-style judgment-last prompt later denoted R5.

\begin{table}[t]
\centering
\caption{Stage~1 backbone screening and dense zero-shot VLM baselines on UCF-Crime. Results are test-set F-AUC/F-AP percentages; $n$ is the number of runs.}
\label{tab:stage1_backbones}
\begin{tabular}{llccc}
\toprule
Approach & Model & $n$ & F-AUC (\%) & F-AP (\%) \\
\midrule
Central MIL & CLIP+MLP & 25 & $83.8 \pm 0.4$ & $23.5 \pm 0.7$ \\
Central MIL & 3D CNN-64 & 2 & $74.3 \pm 0.6$ & $16.6 \pm 1.3$ \\
Central MIL & 3D CNN-112 & 3 & $75.1 \pm 1.5$ & $17.0 \pm 1.0$ \\
\midrule
Fed MIL & CLIP+MLP & 75 & $83.0 \pm 0.6$ & $23.1 \pm 1.2$ \\
Fed MIL & 3D CNN-64 & 1 & $70.6$ & $15.2$ \\
Fed MIL & 3D CNN-112 & 1 & $74.9$ & $17.1$ \\
\midrule
Zero-shot basic & InternVL & 1 & $74.7$ & $18.3$ \\
Zero-shot basic & Qwen3 & 1 & $55.1$ & $8.4$ \\
Zero-shot VERA (R5) & InternVL & 1 & $65.3$ & $10.8$ \\
Zero-shot VERA (R5) & Qwen3 & 1 & $49.2$ & $7.4$ \\
\bottomrule
\end{tabular}
\end{table}

CLIP+MLP outperforms the best 3D CNN by $8.7$ F-AUC points in the centralized setting and $8.1$ points in the federated setting, with the same trend in F-AP. Because the gap was already substantial, we did not expand the CNN baselines into a full multi-seed benchmark. The federated CLIP+MLP scorer incurs only a modest degradation relative to centralized training, with an F-AUC drop of $0.8$ points, and is therefore used as the Stage~1 backbone.

For cascade evaluation, we retain five representative federated CLIP+MLP checkpoints whose F-AUCs range from $83.78\%$ to $84.00\%$, with a mean of $83.86 \pm 0.08$. Since cascade results are reported as changes relative to each checkpoint's MIL baseline, this selection does not bias the comparison between verification regimes. Unless otherwise stated, cascade results are averaged over these five checkpoints.

Dense zero-shot VLM scoring falls below the federated CLIP+MLP baseline. The gap is especially large for Qwen3 basic, which drops by nearly $28$ F-AUC points and labels $88.6\%$ of segments as anomalous, compared with $23.3\%$ for InternVL. VERA also degrades both VLMs in dense zero-shot use, despite later being the strongest cascade prompt for Qwen3 in Table~\ref{tab:textgen_cross_model}. This suggests that MIL routing acts as a cost-saving filter and reduces reliance on unstable dense zero-shot VLM behavior.

This screening is not intended to claim state-of-the-art VAD performance. Recent federated WSVAD methods report higher or comparable UCF-Crime AUC under richer protocols: FedVAD reports $85.13\%$ weakly supervised AUC~\cite{qi2024fedvad}, while FedWSAD reports $84.06\%$ federated AUC on its random split~\cite{wang2025fedwsad}. These results are not directly comparable because they use different client splits and train additional semantic distillation, prompt-learning, or context-driven mechanisms; our focus is the interface between a lightweight federated scorer and frozen VLM verification.

\subsection{Text-Generation Strategy Exploration}
\label{sec:prompt_exploration}

We first analyze the sensitivity of text-generation verification to the prompt strategy. This diagnostic experiment evaluates 14 strategies with InternVL3.5-2B on one high-performing federated Stage~1 checkpoint, using the raw correction rule without temporal post-processing.

The strategies vary along four main axes: direct binary judgment, VERA-style guiding questions, CoT-style reasoning, and score-output prompting. Additional variants inject global context, MIL score context, event-neighborhood context, or a normality prior. Judgment-first and judgment-last variants differ only in whether the final NORMAL/ANOMALOUS decision is requested before or after the explanation. Table~\ref{tab:prompt_exploration} reports raw F-AUC changes before temporal post-processing and is sorted by improvement.

\begin{table}[t]
\centering
\caption{Diagnostic prompt-strategy sweep on one federated Stage~1 checkpoint. Results are raw F-AUC changes relative to the MIL baseline.}
\label{tab:prompt_exploration}
\begin{tabular}{lc}
\toprule
Strategy & $\Delta$F-AUC (pp) \\
\midrule
R13 Score-based output & $+0.254$ \\
R5 VERA, judgment last & $+0.216$ \\
R11 Direct + context & $+0.156$ \\
R2 VERA, judgment first & $+0.077$ \\
R3 CoT, judgment first & $+0.021$ \\
R7 VERA + context & $-0.018$ \\
R4 CoT, judgment last & $-0.027$ \\
R9 VERA + MIL scores & $-0.112$ \\
R1 Direct binary question & $-0.628$ \\
R8 CoT + context & $-0.700$ \\
R14 Debiased VERA + context & $-0.727$ \\
R6 Multi-round CoT & $-1.046$ \\
R12 Debiased VERA & $-1.192$ \\
R10 VERA + event context & $-1.239$ \\
\bottomrule
\end{tabular}
\end{table}

Changing the text-generation strategy provides only limited raw gains. Among binary judgment strategies, only R5 and R11 improve the MIL baseline by more than $0.1$ pp, with most other strategies remaining near or below the baseline. R13 obtains the largest raw gain but is score-based rather than binary. Additional R13 runs show that this numeric output is discretized, with InternVL mostly using scores 3, 5, and 7 rather than the full 0--10 scale. A similar score-collapse behavior appears for Qwen3 in the cross-model study, where R13 collapses to \{5,7\} and is handled with the diagnostic parser adaptation reported in Table~\ref{tab:textgen_cross_model}.

Manual inspection suggests that chain-of-thought responses often mix normal and anomalous cues, making keyword parsing unstable. Strategies that separate the final judgment from the explanation, such as R5, are less prone to this issue. More broadly, only 5 of 14 strategies improve over the MIL baseline, and VERA-style prompts do not transfer uniformly to selective verification despite their motivation from VLM-based anomaly reasoning~\cite{ye2025vera}. These findings motivate evaluating cross-model behavior, temporal post-processing, and direct logit extraction.

\subsection{Cross-Model Text-Generation Verification and Post-Processing}
\label{sec:textgen_postprocessing}

We retain five representative strategies from the diagnostic sweep, spanning high-performing binary prompts, contextual variants, debiased variants, and the score-output prompt. We evaluate them across VLM backbones and the five representative Stage~1 checkpoints, with and without temporal post-processing.

\begin{table}[t]
\centering
\caption{Cross-model text-generation verification. Results are F-AUC changes over five Stage~1 checkpoints; +PP denotes diagnostic F-AUC-selected temporal post-processing.}
\label{tab:textgen_cross_model}
\begin{tabular}{llcc}
\toprule
VLM & Strategy & Raw & +PP \\
\midrule
InternVL & R5  & $+0.268 \pm 0.091$ & $+0.461 \pm 0.177$ \\
InternVL & R7  & $+0.019 \pm 0.105$ & $+1.213 \pm 0.260$ \\
InternVL & R11 & $+0.120 \pm 0.030$ & $\mathbf{+1.553 \pm 0.211}$ \\
InternVL & R12 & $-1.364 \pm 0.165$ & $+1.015 \pm 0.371$ \\
InternVL & R13 & $+0.294 \pm 0.118$ & $+0.684 \pm 0.261$ \\
\midrule
Qwen3 & R5  & $+0.469 \pm 0.093$ & $\mathbf{+1.541 \pm 0.257}$ \\
Qwen3 & R7  & $+0.077 \pm 0.061$ & $+0.435 \pm 0.226$ \\
Qwen3 & R11 & $+0.097 \pm 0.111$ & $+0.590 \pm 0.202$ \\
Qwen3 & R12 & $-1.004 \pm 0.119$ & $+0.330 \pm 0.204$ \\
Qwen3 & R13$^\dagger$ & $+0.245 \pm 0.097$ & $+0.664 \pm 0.211$ \\
\bottomrule
\end{tabular}

\smallskip
\footnotesize
$^\dagger$ Qwen3 R13 collapses to \{5,7\}; 5 is mapped to normal-like and 7 to anomaly-like.
\end{table}

Raw text-generation correction gives modest and inconsistent gains, while temporal post-processing substantially improves several configurations when optimizing for F-AUC. We treat +PP as a diagnostic upper bound on how much temporal regularization can rescue parsed text-generation outputs, not as a deployment-ready model-selection protocol. The strongest text-generation result for InternVL is R11+PP, reaching $+1.553 \pm 0.211$ pp; for Qwen3, R5+PP reaches $+1.541 \pm 0.257$ pp.

The results expose two issues. First, text-output parsing is fragile: Qwen3 R13 requires a model-specific parser adaptation. Second, +PP rankings conflate VLM signal quality with temporal regularization, as shown by InternVL R12 swinging by $+2.4$ pp under post-processing, from $-1.364$ to $+1.015$. Since +PP is selected by F-AUC, its effect on F-AP is not necessarily aligned; we report F-AP only for the best configuration per VLM in Table~\ref{tab:textgen_vs_logit}.

\subsection{Text-Generation versus Logit-Based Verification}
\label{sec:textgen_vs_logit}

We compare the strongest post-processed text-generation configuration for each VLM from Table~\ref{tab:textgen_cross_model} with the main logit configuration, $\lambda=1$, without temporal smoothing. This is the main comparison of the paper. Text-generation results are reported in their most favorable diagnostic setting, including F-AUC-selected post-processing, whereas logit verification uses a fixed rule with no temporal smoothing.

\begin{table}[t]
\centering
\caption{Best text-generation configurations versus logit verification. Results are changes relative to the MIL baseline over five Stage~1 checkpoints.}
\label{tab:textgen_vs_logit}
\begin{tabular}{llcc}
\toprule
VLM & Regime & $\Delta$F-AUC (pp) & $\Delta$F-AP (pp) \\
\midrule
InternVL & Text-gen R11+PP & $\mathbf{+1.553 \pm 0.211}$ & $+1.051 \pm 0.469$ \\
InternVL & Logit, $\lambda=1$ & $+1.053 \pm 0.086$ & $\mathbf{+4.638 \pm 0.375}$ \\
\midrule
Qwen3 & Text-gen R5+PP & $\mathbf{+1.541 \pm 0.257}$ & $+1.551 \pm 0.435$ \\
Qwen3 & Logit, $\lambda=1$ & $+1.522 \pm 0.102$ & $\mathbf{+7.721 \pm 0.637}$ \\
\bottomrule
\end{tabular}
\end{table}

Post-processed text generation can be competitive in F-AUC but has higher variance and weaker F-AP gains. Using the mean Stage~1 baseline of $83.86\%$, the main logit configuration corresponds to approximately $84.91\%$ F-AUC with InternVL and $85.38\%$ with Qwen3. These absolute values should not be read as direct comparisons to FedVAD or FedWSAD because the protocols and client splits differ, but they show that selective VLM verification can lift a lightweight federated scorer into a competitive F-AUC range. More importantly, the logit regime yields substantially larger F-AP improvements: $4.4\times$ larger for InternVL and $5.0\times$ larger for Qwen3 than the best text-generation configuration. This is consistent with the continuous logit signal preserving rank information across suspect segments, while parsed text-generation corrections can map normal-corrected segments through the same attenuation rule, $s_i=0.1m_i$, compressing the score distribution among them.

The main advantage of the logit interface is therefore not peak F-AUC under tuned post-processing, but interface robustness. In this selective MIL--VLM cascade, next-token logits avoid parser failures, preserve a continuous ordering among suspect segments, improve F-AP substantially, and remain causal without temporal smoothing. They should be viewed as a more deployable verification signal, not as evidence that VLM logits are universally better calibrated than generated text.

\subsection{Causal versus Bidirectional Evaluation}
\label{sec:causal_bidirectional}

We finally evaluate causal compatibility for streaming-oriented surveillance (Table~\ref{tab:causal_bidirectional}). We use the term causal in the segment-scoring sense: once a suspect segment and its sampled frames are available, its VLM score does not depend on future segments. This does not model all constraints of a fully online streaming system, such as unknown video duration or adaptive segmentation.

\begin{table}[t]
\centering
\caption{Causal versus bidirectional F-AUC changes. Logit rows use $\lambda=1$ without temporal smoothing, so bidirectional and causal scores are identical in the segment-scoring sense.}
\label{tab:causal_bidirectional}
\scriptsize
\begin{tabular}{lcc}
\toprule
Configuration & Bidir. & Causal \\
\midrule
InternVL R5+PP  & $+0.461 \pm 0.177$ & $+0.359 \pm 0.078$ \\
InternVL R11+PP & $+1.553 \pm 0.211$ & $+1.116 \pm 0.189$ \\
InternVL R12+PP & $+1.015 \pm 0.371$ & $+0.478 \pm 0.236$ \\
InternVL R13+PP & $+0.684 \pm 0.261$ & $+0.348 \pm 0.115$ \\
Qwen3 R5+PP     & $+1.541 \pm 0.257$ & $+1.019 \pm 0.183$ \\
Qwen3 R7+PP     & $+0.435 \pm 0.226$ & $+0.156 \pm 0.082$ \\
Qwen3 R11+PP    & $+0.590 \pm 0.202$ & $+0.304 \pm 0.083$ \\
Qwen3 R12+PP    & $+0.330 \pm 0.204$ & $+0.055 \pm 0.072$ \\
Qwen3 R13$^\dagger$+PP & $+0.664 \pm 0.211$ & $+0.359 \pm 0.085$ \\
\midrule
InternVL logit, $\lambda=1$ & $+1.053 \pm 0.086$ & $+1.053 \pm 0.086$ \\
Qwen3 logit, $\lambda=1$ & $+1.522 \pm 0.102$ & $+1.522 \pm 0.102$ \\
\bottomrule
\end{tabular}
\end{table}

All post-processed text-generation configurations lose F-AUC under the causal constraint, with drops of $0.10$ to $0.54$ pp. For the logit verifier, the bidirectional and causal columns are identical by construction because $\lambda=1$ substitutes each suspect segment with its own VLM logit score and applies no temporal smoothing. The logit interface is therefore causal-compatible at the segment-scoring level, unlike post-processed text-generation verification.

\section{Conclusion}

We presented a lightweight federated MIL--VLM cascade for weakly supervised video anomaly detection. Only a compact MIL head is trained federatively on frozen CLIP features, while a frozen VLM is used post hoc to verify selected suspect segments and remains decoupled from federated optimization.

Experiments show that text-generation verification is sensitive to prompts, parsing, model choice, and smoothing. In contrast, the parser-free logit interface improves both F-AUC and F-AP over the MIL baseline across both VLMs without temporal smoothing, and is causal at the segment-scoring level when $\lambda=1$.

Limitations include centralized offline VLM evaluation, validation only on UCF-Crime, four physical CPU-only clients with Dirichlet-partitioned data rather than naturally occurring deployment silos, pre-extracted CLIP features, and fixed routing/fusion settings. Future work should test ShanghaiTech, XD-Violence, broader VLM families, additional federated partitions, and deployment-specific privacy/latency constraints.

\end{document}